\documentclass{bmcart}

\usepackage[utf8]{inputenc} %unicode support
\usepackage{verbatim}
\usepackage{graphicx}
\usepackage{epstopdf}
\usepackage{multirow}
\usepackage{textcomp}
\usepackage{adjustbox}

\startlocaldefs
\endlocaldefs

\begin{document}

%%% Start of article front matter
\begin{frontmatter}

\begin{fmbox}
\dochead{Research}


\title{Feature versus Raw Sequence: Deep Learning Comparative Study on Predicting Pre-miRNA}

%%%%%%%%%%%%%%%%%%%%%%%%%%%%%%%%%%%%%%%%%%%%%%
%%                                          %%
%% Enter the authors here                   %%
%%                                          %%
%% Specify information, if available,       %%
%% in the form:                             %%
%%   <key>={<id1>,<id2>}                    %%
%%   <key>=                                 %%
%% Comment or delete the keys which are     %%
%% not used. Repeat \author command as much %%
%% as required.                             %%

\author[
   addressref={aff1},                   % id's of addresses, e.g. {aff1,aff2}
   noteref={n1},                        % id's of article notes, if any
   email={jaya.thomas@sunykorea.ac.kr}   % email address
]{\inits{JT}\fnm{Jaya} \snm{Thomas}}
\author[
   addressref={aff1},
   noteref={n1},                        % id's of article notes, if any
   email={sonia.thomas@sunykorea.ac.kr}
]{\inits{ST}\fnm{Sonia} \snm{Thomas}}
\author[
   addressref={aff1,aff2},
   corref={aff1,aff2},    % id of corresponding address, if any
   email={sael@cs.stonybrook.edu}
]{\inits{LS}\fnm{Lee} \snm{Sael}}
%%%%%%%%%%%%%%%%%%%%%%%%%%%%%%%%%%%%%%%%%%%%%%
%%                                          %%
%% Enter the authors' addresses here        %%
%%                                          %%
%% Repeat \address commands as much as      %%
%% required.                                %%
%%                                          %%
%%%%%%%%%%%%%%%%%%%%%%%%%%%%%%%%%%%%%%%%%%%%%%

\address[id=aff1]{%                           % unique id
  \orgname{Department of Computer Science, SUNY Korea}, % university, etc
  %\street{},                     %
  \postcode{21985}                                % post or zip code
  \city{Incheon},                              % city
  \cny{Korea}                                    % country
}
\address[id=aff2]{%
  \orgname{Department of Computer Science, Stony Brook University},
  \postcode{11794}
  \city{Stony Brook, NY},
  \cny{USA}
}

%%%%%%%%%%%%%%%%%%%%%%%%%%%%%%%%%%%%%%%%%%%%%%
%%                                          %%
%% Enter short notes here                   %%
%%                                          %%
%% Short notes will be after addresses      %%
%% on first page.                           %%
%%                                          %%
%%%%%%%%%%%%%%%%%%%%%%%%%%%%%%%%%%%%%%%%%%%%%%

\begin{artnotes}
%\note{Sample of title note}     % note to the article
\note[id=n1]{Equal contributor} % note, connected to author
\end{artnotes}

\end{fmbox}% comment this for two column layout

%%%%%%%%%%%%%%%%%%%%%%%%%%%%%%%%%%%%%%%%%%%%%%
%%                                          %%
%% The Abstract begins here                 %%
%%                                          %%
%% Please refer to the Instructions for     %%
%% authors on http://www.biomedcentral.com  %%
%% and include the section headings         %%
%% accordingly for your article type.       %%
%%                                          %%
%%%%%%%%%%%%%%%%%%%%%%%%%%%%%%%%%%%%%%%%%%%%%%

\begin{abstractbox}

\begin{abstract} % abstract
\parttitle{Background}
Should we input known genome sequence features or input sequence itself in deep learning framework? As deep learning more popular in various applications, researchers often come to question whether to generate features or use raw sequences for deep learning.
To answer this question, we study the prediction accuracy of precursor miRNA prediction of feature-based deep belief network and sequence-based convolution neural network.
\parttitle{Results}
Tested on a variant of six-layer convolution neural net and three-layer deep belief network, we find the raw sequence input based convolution neural network model performs similar or slightly better than feature based deep belief networks with best accuracy values of 0.995 and 0.990, respectively.
Both the models outperform existing benchmarks models.
The results shows us that if provided large enough data, well devised raw sequence based deep learning models can replace feature based deep learning models. However, construction of well behaved deep learning model can be very challenging. In cased features can be easily extracted, feature-based deep learning models may be a better alternative.
\end{abstract}

%%%%%%%%%%%%%%%%%%%%%%%%%%%%%%%%%%%%%%%%%%%%%%
%%                                          %%
%% The keywords begin here                  %%
%%                                          %%
%% Put each keyword in separate \kwd{}.     %%
%%                                          %%
%%%%%%%%%%%%%%%%%%%%%%%%%%%%%%%%%%%%%%%%%%%%%%


% MSC classifications codes, if any
%\begin{keyword}[class=AMS]
%\kwd[Primary ]{}
%\kwd{}
%\kwd[; secondary ]{}
%\end{keyword}

\end{abstractbox}
%
%\end{fmbox}% uncomment this for twcolumn layout

\end{frontmatter}

%%%%%%%%%%%%%%%%%%%%%%%%%%%%%%%%%%%%%%%%%%%%%%
%%                                          %%
%% The Main Body begins here                %%
%%                                          %%
%% Please refer to the instructions for     %%
%% authors on:                              %%
%% http://www.biomedcentral.com/info/authors%%
%% and include the section headings         %%
%% accordingly for your article type.       %%
%%                                          %%
%% See the Results and Discussion section   %%
%% for details on how to create sub-sections%%
%%                                          %%
%% use \cite{...} to cite references        %%
%%  \cite{koon} and                         %%
%%  \cite{oreg,khar,zvai,xjon,schn,pond}    %%
%%  \nocite{smith,marg,hunn,advi,koha,mouse}%%
%%                                          %%
%%%%%%%%%%%%%%%%%%%%%%%%%%%%%%%%%%%%%%%%%%%%%%

%%%%%%%%%%%%%%%%%%%%%%%%% start of article main body
% <put your article body there>

%%%%%%%%%%%%%%%%
%% Background %%
%%
\section*{Introduction}
Deep learning methods have been popularized in bio-sequence analysis.
More specifically, convolution neural network (CNN) have been widely applied to characterize and classify raw sequence data \cite{}.
Traditionally, to classify sequence data, sequence features were generated, fed in to classification algorithms, and predictions were made.
CNN simplified this process by removing the need to feature generation which could be challenging in cases.
However, question arises of whether to use raw sequences when there are already good set of features.
We answer this question in the context of precursor micro RNA prediction where raw sequence data as well as set of working sequence features which are shown to give high accuracy.

\subsection*{Precursor miRNA}
MicroRNAs (miRNAs) are single-stranded small non-coding RNAs that are typically 22 nucleotides long.
A miRNA regulates gene expression at the post transcription level by base pairing with a complementary messenger RNA (mRNA) there by hindering the translation of the mRNA to proteins.
The regulatory role of miRNAs are important in development, cell proliferation and cell death and their malfunction has been connected with neuro-degenerative disease, cancer and metabolic disorders \cite{Witkos2011}.
Furthermore, informatics analysis predicts that 30$\%$ of human genes are regulated by miRNA~\cite{Ross2007}.

MiRNAs can be experimentally determined by directional cloning of endogenous small RNAs \cite{Chen2005}.
However, this is a time consuming process that require expensive laboratory reagents.
These drawbacks motivate the application of computational approaches for predicting miRNAs.
The goal of miRNA prediction is to correctly classify pre-miRNAs from other pseudo hairpins.
Via miRNA biogenesis, pre-miRNA becomes a mature miRNA, however, other hairpins do not.
The miRNA biogenesis involves number of steps.
First, primary transcripts of miRNA (pri-miRNA) are transcribed from introns of protein coding genes that are several kilobases long.
The pri-miRNAs are then clopped by Rnase-III enzyme Drosha into $\sim$70 base pairs (bp) long hair-pin-looped precursor miRNAs (pre-miRNAs).
Then exportin-5 proteins transport pre-miRNA hairpins into the cytoplasm through nuclear pore.
In cytoplasm, pre-miRNAs are further cleaved by Rnase-III enzyme Dicer to produce a $\sim$20 bp double stranded intermediate called miRNA:miRNA*.
Then a strand of the duplex with the low thermodynamic energy becomes a mature miRNA.
% Most mature miRNAs interact with a RNA interference (RNAi) induced silencing complex (RISC) through base pairing of the target mRNAs and regulate the expression of the genes.
\subsection*{Precursor miRNA prediction methods}
Several machine learning based methods have been proposed to predict miRNAs, that is to determine the true pre-miRNAs from other pseudo hairpins, RNA sequences that have similar stem-loop features to pre-miRNAs but does not contain mature miRNAs,  with high accuracy.
Most methods relies on features generated from  sequence, folding measures, stem-loop features and statistical measures and careful selection of features.

Many tools have been developed based on the different classification techniques such as naive Bayes classifier (NBC), artificial neural networks (ANN), support vector machines (SVM), and random forests (RF).
Among the approaches, support vector machine (SVM) had been most extensively applied.
Some of the notable SVM-based methods includes triplet-SVM~\cite{Xue2005}, MiRFinder~\cite{Huang2007}, miPred~\cite{Ng2007}, microPred~\cite{Batuwita2009}, yasMiR~\cite{Pasaila2011}, MiRenSVM~\cite{Ding2010}, MiRPara~\cite{Wu2011}, YamiPred~\cite{Kleftogiannis2015}, and G$^{2}$DE \cite{Hsieh2010}.
Among them, Triplet-SVM\cite{Xue2005} is the classifier that consider local structure-sequence features that reflect characteristics of miRNAs.
The approach report an accuracy of 90$\%$ considering pre-miRNAs from the other 11 species including plants and virus without considering any other comparative genomics information.
Another, miPred\cite{Ng2007} SVM approach considered Gaussian Radial Basis Function kernel (RBF) as a similarity measure for  global and intrinsic hairpin folding attribute and resulted with accuracy of around 93$\%$.
MicroPred\cite{Batuwita2009} introduces some additional features for evaluation of miRNA using SVM based machine learning classifier. Author's report classification results of high specificity of 97.28$\%$ and sensitivity of 90.02$\%$.
The miR-SF classifier \cite{Wang2011} predicts the identified human pre-miRNAs in miRBase source on the selected optimized feature subset including 13 features, generated by SVM and genetic algorithm.
Finially, YamiPred \cite{Kleftogiannis2015} is a genetic algorithm and SVM based embedded classifier that consider feature selection and parameters optimization for improving performance.
Other notable methods are based on random forests (RF)\cite{Xiao2011} and artificial neural networks (ANN)~\cite{Rahman2012, Thomas2017}.
In MiRANN\cite{Rahman2012} predictor, author's consider neural network for pre-miRNA prediction by expanding the network with more neurons and the hidden layers and reports an $99.9\%$ ACC on a human dataset.
The network is designed to be impartial for any feature by integrating exceptional weight initializing equation where closest neurons slightly differ in weights.
MiRANN utilizes carefully selected features on a neural network structure.
However, to the best of our knowledge, raw sequence data have not been used for distinguish pre-miRNA from other hair-pin sequences.

\subsection*{Deep learning approaches}
Two types of neural network models, i.e., deep belief network and convolution neural network, are used to to compare the prediction accuracy of feature-based learning and raw sequence based learning.
Convolution neural network (CNN) has been used in several instances to directly process raw data as input.
CNN has gained momentum due to its success in improving the previously recorded state-of-the-art performance measures in a wide range of domains including genome sequence analysis.
Deep belief network (DBN), on the other hand has been popular with where there are large number of features.
Whether the input is a raw data or a high dimensional feature, DNN uses multi-layer architecture to infer from data.
The deep architecture automatically extracts high-level feature necessary for classification or clustering.
That is, the multiple layers in deep learning helps exploit the inherent complexities of data.

\subsubsection*{Deep belief network} \label{sec:DBN}
Deep belief network (DBN) is an architecture obtained by stacking multiple restricted boltzmann machines (RBMs), such that the $i^{th}$ hidden layer is the input to the $i+1^{th}$ hidden layer.
Let $X$ be the observed vector and hidden layer $H_k$, with N hidden layers, then the distribution is as follows \cite{Hinton2002}:
\begin{equation}
P(X,H^1,H^2,...,H^k)=P(H^{N-1},H^N)(\prod_{k=0}^{N-2}P(H^k|H^{k+1})),
\end{equation}
where $X=H^0$, $P(H^{k-1)}|H^k)$ is the distribution for visible (input) unit on the $k^{th}$ level hidden layer, and $P(H^{k-1}|H^k)$ is the top level layer distribution of the visible-hidden layer.
The first step in training the DBN is to train the first layer (visible layer) of the model such that is models the raw input $X=H0$.
In the second step the distribution of the input (i.e transformed data) is obtained as $P(H^1|H^0)$ using training results  of the first layer and is used as the input for the second layer.
In the third step, second layer of RBM is trained on sampling the learned conditional probability in the previous layer.
Steps two and three is repeated to generate multiple layers.
In the final step, hyper parameters are fine-tuned based on the gradient descent based back propagations.
The first hidden layer learns from the structure of the data through the input layer and the process is continued by adding the second layer.
The first hidden layer acts as the input, which is multiplied by weight at the nodes of second hidden layer and thus the probability for activating the second hidden layers is calculated.
This process results in sequential sets of activations by grouping features of features resulting in a feature hierarchy, by which networks learn more complex and abstract representations of data, and can be repeated several times to create a multi-layer network.
A standard feed-forward neural network is added after the last hidden layer to predict the label, the input to the network being being the activation probabilities.
The resulting DBN is put together to adjust the weights with stochastic gradient descent back propagation \cite{Hinton2012}.

\subsubsection*{Convolution neural network} \label{sec:CNN}
Typical CNN models consist of multiple layers of convolution layer and pooling layer alternations finalized by a fully connected layer.
The convolution layer performs the convolution operation between the input values and learned filters, matrix of weights.
Let $(m,n)$ be the filter size and $W$ be the small matrix of weights, then the convolution layer performs a modified convolution of the W with the input X by calculating the dot product $W\cdot x+b$, where $x$ a instance of $X$ and $b$ is the bias.
Typically the filters are are share by using the same filter across different positions of the input.
The step size by which the filter slides across the input is called the stride, and the filter area $(m \times n)$ is called the receptive field.
Weight sharing concept is the important characteristics of a convolution network that reduces the complexity of the problem by reducing number of weights learned.
It is also allows location invariant learning, i.e., if a important pattern exists, a appropriate CNN model will learn it no matter where in the sequence.
The convolution layer is often followed by the pooling layer that summaries the value learned in the convolution layer.
The pooling also allows invariance in the learning as well as reducing model complexity.
Popular pooling methods are average pooling or max pooling.
The final layer is the fully connected layer which is connected to the output or the classification layer

%The activation map is the dot product at each position, i.e, they are the partially connected neurons having same weights.
%This layer basically operates on the activation map and tries to reduce its dimension by grouping.

\subsection*{Contributions}
 The main contribution of the paper are summarized as follows:
\begin{itemize}
  \item Compare the performance of feature based and raw sequence based deep learning: a deep belief network models is proposed for integrating large number of heterogeneous features and convolution neural network model is proposed for the raw input sequence data.
  \item Provides a solution for class imbalance problem, allowing for unbiased performance measures.
  \item Compares the performance of proposed model against existing machine learning classifier on eleven different species which extends the previous work on human dataset only \cite{Thomas2017}.
\end{itemize}

\section*{Methods}
\subsection*{Dataset}
\subsubsection*{miRNA data selection}
We use experimentally validated pre-miRNAs as positive examples and pseudo hairpins as negative examples to train and test the proposed method.
The human pre-miRNA sequence was retrieved from the miRBase 18.0 release~\cite{Zhong2015}.
Similar to miPred \cite{Ng2007} approach, the multiple loops were discarded to obtain 1600 pre-miRNAs (positive) dataset.
The positive sample sequence has an average length of 84 nt with maximum and minimum length being 154 nt and 43 nt respectively. Similarly the negative sample sequence has average length of 85nt with 63 nt and 120 nt as minimum and maximum length respectively.
The negative dataset consists of 8494 pseudo hairpins as the false samples.
They were extracted from the human protein-coding regions as suggested by microPred \cite{Batuwita2009}.
The average length of the sequence is 85 nt with minimum as 63 nt and maximum as 120 nt.  The different filtering criteria, including non-overlapping sliding window, no multiple loops, lowest base pair number set to 18, and minimum free energy less than 15kcal/mol were applied on these sequences to resemble the real pre-miRNA properties.

\begin{comment}
To evaluate the performance of the proposed model, we also consider miRNA data for 10 other species. Some of these species are similar to human including gorillas and chimpanzees, where as others are distinct including Aves and Rodentia. The number of positive and negative data samples for considered species are summarized in Table ~\ref{table:00}. A more detailed description on the dataset can be refered here \cite{Kleftogiannis2015}.

The training set (TE-H) were randomly selected 200 human pre-miRNAs from miRBase 8.2, and 400 pseudo-miRNA considered in TripleSVM.
Here, TE-H dataset contains 123 human  pre-miRNAs remaining from miRBase 8.2 after allocation of pre-miRNA for training (TR-H).
The IE-NH dataset included 1918 pre-miRNA from 40 non-human species from miRBAse 8.2, IE-NC consist of 12387 non-coding RNAs from Rfam 7.0 database \cite{Griffiths-Jones2005}, and IE-M included 31 messenger RNAs selected from GenBank \cite{Benson2007}.
\end{comment}

\subsubsection*{Class imbalance solution}
Another problem that we have addressed here is the class imbalance problem in miRNA predictions.
Class imbalance is a machine learning problem where the number of data samples belonging to one class (positive) is far less compared to data sample belonging to other class (negative).
The imbalance class is often solved by using either under or over sampling methods. In case of under sampling the data samples are removed from the majority class, whereas for over sampling balance is created either by addition of new samples or duplication of the existing minority class samples.
Class imbalance problem often arise in miRNA data classification problem due to abundance in pseudo hairpin structure compared to true pre-miRNAs folds.
In existing classifiers such as triplet-SVM\cite{Xue2005} and  miPred \cite{Ng2007} handled the imbalance problem manually.

We address the class imbalance problem during the training phase by adopting a modified under sampling approach \cite{Yen2009}. In the modified approach, we divided the majority class into subsets using k-means algorithm with k=5, and thus obtain clusters with slightly higher similarity amongst the group.
The entire negative samples is divided into subsets using k-means algorithm with k=5, and the cluster having the highest similarity index among the group is selected.
Now using 8 fold cross validation, the negative samples is divided into training and testing dataset such that training dataset has 1400 negative samples and testing dataset has 200 negative samples.
Similarly, the positive sample is divided into training and testing dataset using 8 fold cross validation such that training dataset has 1400 positive sample and testing dataset has 200 positive samples.
Hence, the training dataset has 2400 samples and testing dataset has 400 samples.

% CHANGE
\subsection*{Modeling deep belief network}
\subsubsection*{miRNA feature encoding}
Feature based learning require features as inputs.
This work adopts 58 characteristic features, which are shown useful in existing studies for predicting miRNA \cite{Kleftogiannis2015}.
The features includes sequences composition properties, folding measures, stem-loop features, energy and statistical features, and 20 selected features to differentiate pre-miRNAs from pseudo hairpins for candidate input of the DBN model.
These features are extracted based on the knowledge based analysis of the existing methods for the miRNA analysis.
The common characteristics of pre-miRNAs used for evaluation consists of sequences composition properties, secondary structures, folding measures and energy.
The sequence characteristics include features related to the frequency of two and three adjacent nucleotide and aggregate dinucleotide frequency in the sequence.
The secondary structure features from the perspectives of miRNA bio-genesis relating different thermodynamic stability profiles of pre-miRNAs.
These structures have lower free energy and often contain stem and loop regions.
They include diversity, frequency, entropy-related properties, enthalpy-related properties of the structure.
The other features are hairpin length, loop length, consecutive base-pairs and ratio of loop length to hairpin length of pre-miRNA secondary structure.
The energy characteristic associated to the energy of secondary structure includes the minimal free energy of the secondary structure, overall free energy NEFE, combined energy features and the energy required for dissolving the secondary structure.
All the features extracted are normalized to standardizing the inputs in order to improve the training and to avoid getting stuck in local optima. The features used  are summarized Table~\ref{tab:DBN_feat} and detailed in Table~\ref{tab:S1} .

\subsubsection*{Deep belief network architecture}

{
\begin{figure}[!ht]
  \centering
  \caption{A deep learning to predict miRNA with extracted features.}
\label{fig:01}
\end{figure}
}

The proposed DBN based miRNA prediction method, we call miRNA-FDL, has three hidden layers, and the model is denoted as  X-100-70-35-1, where X being the size of the input layer, 1 denotes the number of neuron in the output layer and the remaining values denotes the number of neurons in each hidden layer.
Figure \ref{fig:01} illustrates the model architecture and layer-by-layer learning procedure described in \label{sec:DBN}.
Different model architectures were trained using the same learning procedure but varying the number of hidden layer and nodes.
Amongst the candidate network models, a better one was selected based on the classifier accuracy.

The weights of the miRNA-FDL was trained with stochastic gradient descent base back propagation algorithm \cite{Hinton2012}] were the update rule is the following:
\begin{equation}
w_{ij}(t+1)=w_{ij}(t) + \eta \frac{\partial C}{\partial w_{ij}},
\end{equation}
where, $ w_{ij} (t+1)$ is the weight computed at $t+1$,  $\partial$ denotes the learning rate, and $C$ is the cost function.  For the given model, softmax is used as an activation function and the cost is computed using cross entropy.
The softmax function is defined as
\begin{equation}
p_{j}=\frac{exp(x_{j})}{ \sum_{k}exp(x_{k})},
\end{equation}
where, $p_{j}$ is the output of the unit j, $x_{j}$ and $x_{k}$ denotes the total input to unit $j$ and $k$, respectively for the same level. The cross entropy is given by
\begin{equation}
C=-\sum_{i} d_{j} log(p_{j}),
\end{equation}
where $d_j$ is the target probability for output unit $j$ and $p_j$ is the probability output after applying the activation function.

\subsection*{Modeling convolution neural network}
\subsubsection*{CNN input processing}
Each pre-miRNA is a RNA sequence of composed letters (A, C, G, U).
Each of the nucleotide is encoded using one-hot-code methods. That is, A is encoded as (1,0,0,0), C as (0,1,0,0), G as (0,0,1,0), and U as (0,0,0,1).
The micro RNA is a nucleotide sequence of (A, C, G, U).
\subsubsection*{Convolution neural network architecture}
%Raw sequence based learning with CNN is trained and tested on the experimentally validated pre-miRNAs sequences as positive examples and pseudo hairpin sequences as negative examples.
Various architecture of the CNN can be generated dependent on the choice of number of layers and on how to combine convolution layer with pooling layers.
The Table \ref{tab:CNN_arc} shows the various variants of the CNN architecture considered for the study.
In CNN model type 1, the CNN architecture has single layer of convolution followed by a pooling layer which in turn is connected to the fully connected output layer.
The output layer is connected to the classification layer which classifies the predicted labels.
The model type 2 is variation to model type 1, such that the pooling layer of model type 1 is replaced by a fully connected layer.
Hence the model type 2 has two fully connected layers.
The further variation to the model type 1 leads to the model type 3 such that it has two convolution layers.
All other layers as similar to the base model 1.
The model type 4 has three convolution layers. In all the models global pooling is preferred over local pooling as it is observed that features are better learned in global pooling.

The architecture of the CNN model highly depends on the various hyper-parameters.
We set the number of node in the input layer to be $4 \times 160$, 4 for the one-hot-code encoding and 160 the for length of the sequence as the maximum length of sequence in the database is 154bp.
We set the output node of the the fully connected layer to be three and add a classification layer which identifies the input sequence as whether it contains the pre-miRNA or not based on the three nodes result.
Other hyper-parameters tested are list in the  Table~\ref{tab:CNN_hp}.
The various combination of the hyper-parameters mentioned in Table~\ref{tab:CNN_hp} with the models mentioned in Table~\ref{tab:CNN_arc}, are considered.

%\subsection*{Performance Measure}
%The efficacy of the proposed pre-miRNA prediction model are based on the following performance metrics. True positive (TP) denotes the number of data samples classified as positive (true pre-miRNAs) and  negative (TN) represent correctly classified negative samples (pseudo pre-miRNAs). Similarly false positive (FP) and false negative (FN) represents the numbers of the misclassified positive and negative samples, respectively.

\section*{Results}
Wether to determine the efficacy of raw sequence based learning and feature based learning, we compared the accuracies of two DBN models that works one unselected pre-miRNA feature set of size 58 and selected feature set of size 20 and the accuracy of one CNN model that works on raw pre-miRNA sequences.
We also compared the proposed methods on other machine learning methods.
The proposed miRNA prediction models are implemented on MATLAB 2016 (b) platform with 2.30 GHz Intel Xenon GPU E5-2630 and 32 GB RAM.
The most crucial aspect of the deep learning was on the selection of the appropriate hyper-parameters.
We describe the final models that was selected in the following. The performance of the proposed and compared methods are summarized in Table~\ref{tab:acc}.

\subsection*{DBN based precursor miRNA prediction model}
The various candidate model for the DBN based precursor miRNA prediction model is obtained by varying the number of hidden nodes in the hidden layer as well as the number of hidden layers. The best prediction accuracy is obtained for a DBN network architecture [Fig.~\ref{fig:dbn}] of three hidden layers with first, second and third layer having 100, 70, 35 hidden neurons respectively.
Considering the stochastic nature of the algorithm the output values are averaged for twenty executions.
It is observed that for 58 features as input, the DBN model (Input-100-70-35-output) gives an accuracy of 0.968 with F1-score of 0.957
Furthermore from the literature survey \cite{Kleftogiannis2015} it is learned that the most relevant features associated with the miRNA are melting temperature, enthropy, enthalpy and free minimum energy.
Thus the 20 relevant features mentioned are aggregate nucleotide frequency A+U, dinucleotide frequencies AG, AU, CU, GA, UU, Minimum Free Energy Index 4 (MFEI4), Positional Entropy, Normalized Ensem-ble free energy, Frequency of the MFE structure (Freq), Enthalpy normalized by the length of the sequence (dH/L), Melting temperature (Tm), Melting temperature normalized by length (Tm/L), Normalized base-pair count by length, j G-Cj /L, Normalized average base pairs by number of stem loops (A-U)/ stems, (G-U)/stems, the length of the sequence (Len), Centroid energy normalized by length (CE/L), and the Statistical Z-scores zG and zSP.
The DBN model for the above 20 features gives an accuracy of 0.992 with F score 0.989 which is slightly higher than the using all 58 features.

\subsection*{CNN based precursor miRNA prediction model}
The various candidate models obtained by the combination of Table~\ref{tab:CNN_arc} and Table~\ref{tab:CNN_hp} are bested and two models that output highest accuracies on the validation data set are selected.
Deeper layers were also tested, however, additional layers of convolution does not increases the performance of the miRNA prediction due to two factors due to limitation of available number of data.
The two models are described bellow and summarized in Table~\ref{tab:CNN_best}.

The architecture of the model type 2 could be explained as follows: the input layer (raw sequence data) is convoluted with a filter (window) of size of 18, and the window is shift-ed with value of 4 (using a stride of 4). The total number of filters used are 20. The output obtained after the convolution, is now fed into a fully connected layer having 90 neurons. Furthermore the output from this fully connected layer is again fed into another fully connected layer having 2 neurons. This layer is also called the output layer. The output layer is connected to a classification layer which classifies the label. In the model type 2, after the convolution layer, two fully connected layers are used before the final (output) classification layer. The fully connected layer helps in better learn-ing of the features that are extracted from the convolution layer.

In the model type 3 as depicted by Figure~\ref{fig:cnn}, the architecture is as follows: the input layer (raw se-quence data) is convoluted with a filter (window) of size 12 and the window is shifted with value 1 (Stride =1). The output of this convolution layer is again convoluted with another filter (window) of size 6 and Stride=1. For both the convolutions the number filters used is 12. Now after the second convolution, a max pooling layer is connected with window size 6 and the window is shifted with value 4 (stride =4).  The output of the pooling layer is connected to the fully connected layer having 2 neurons, (i.e the output layer). The output layer is connected to a classification layer which classifies the label. For both the models, the accuracy is at its best at the dropout ratio of 0.3 at the output layer.

\subsection*{Comparison with the existing computational methods}
Proposed prediction model using CNN and DBN are compared to the existing benchmark models.
It is clearly observed that the prediction model based on deep learning approaches outperforms compared methods.
Two models of CNN and DBN model with 20 selected features has highest accuracy values above 0.99. The DBN model working on 58 features also has high accuracy of 0.968. This shows that DBN performs well on large set of unselected features.
Both the proposed models in this study, provided enough data, validate that deeper the network model, higher is the precision efficacy of the classifier.
Table~\ref{tab:acc}.

\section*{Discussions and Conclusions}
In this study, prediction model for prediction of precursor miRNA that contains miRNA sequence is proposed using deep learning techniques using convolution neural networks on raw sequence input and deep belief networks on feature sets.
Convolution neural network, when well modeled, were able to automatically learn relevant feature from raw RNA sequence for predicting correct pre-miRNAs, hence developing a highly accurate classifier.
Deep learning framework, outperform all the existing popular learning algorithms including naive Bayes, random forest, k nearest neighbor, and SVM.

%%%%%%%%%%%%%%%%%%%%%%%%%%%%%%%%%%%%%%%%%%%%%%
%%                                          %%
%% Backmatter begins here                   %%
%%                                          %%
%%%%%%%%%%%%%%%%%%%%%%%%%%%%%%%%%%%%%%%%%%%%%%

\begin{backmatter}

\begin{comment}
\section*{Competing interests}
The authors declare that they have no competing interests.

\section*{Author's contributions}
Text for this section \ldots

\section*{Acknowledgements}
This research was supported by Basic Science Research Program through the NRF of Korea (NRF-2015R1C1A2A01055739), by the KEIT Korea under the “Global Advanced Technology Center” (10053204) and by the MSIP, Korea, under the ”ICT Consilience Creative Program” (IITP-2015-R0346-15-1007) supervised by the IITP.
\end{comment}
%%  After submission of the .TEX file,                     %%
%%  you will be prompted to submit your .BBL file.         %%

% if your bibliography is in bibtex format, use those commands:
\bibliographystyle{bmc-mathphys} % Style BST file (bmc-mathphys, vancouver, spbasic).
\bibliography{mybibfile}      % Bibliography file (usually '*.bib' )


\begin{thebibliography}{20}
\ifx \bisbn   \undefined \def \bisbn  #1{ISBN #1}\fi
\ifx \binits  \undefined \def \binits#1{#1}\fi
\ifx \bauthor  \undefined \def \bauthor#1{#1}\fi
\ifx \batitle  \undefined \def \batitle#1{#1}\fi
\ifx \bjtitle  \undefined \def \bjtitle#1{#1}\fi
\ifx \bvolume  \undefined \def \bvolume#1{\textbf{#1}}\fi
\ifx \byear  \undefined \def \byear#1{#1}\fi
\ifx \bissue  \undefined \def \bissue#1{#1}\fi
\ifx \bfpage  \undefined \def \bfpage#1{#1}\fi
\ifx \blpage  \undefined \def \blpage #1{#1}\fi
\ifx \burl  \undefined \def \burl#1{\textsf{#1}}\fi
\ifx \doiurl  \undefined \def \doiurl#1{\textsf{#1}}\fi
\ifx \betal  \undefined \def \betal{\textit{et al.}}\fi
\ifx \binstitute  \undefined \def \binstitute#1{#1}\fi
\ifx \binstitutionaled  \undefined \def \binstitutionaled#1{#1}\fi
\ifx \bctitle  \undefined \def \bctitle#1{#1}\fi
\ifx \beditor  \undefined \def \beditor#1{#1}\fi
\ifx \bpublisher  \undefined \def \bpublisher#1{#1}\fi
\ifx \bbtitle  \undefined \def \bbtitle#1{#1}\fi
\ifx \bedition  \undefined \def \bedition#1{#1}\fi
\ifx \bseriesno  \undefined \def \bseriesno#1{#1}\fi
\ifx \blocation  \undefined \def \blocation#1{#1}\fi
\ifx \bsertitle  \undefined \def \bsertitle#1{#1}\fi
\ifx \bsnm \undefined \def \bsnm#1{#1}\fi
\ifx \bsuffix \undefined \def \bsuffix#1{#1}\fi
\ifx \bparticle \undefined \def \bparticle#1{#1}\fi
\ifx \barticle \undefined \def \barticle#1{#1}\fi
\ifx \bconfdate \undefined \def \bconfdate #1{#1}\fi
\ifx \botherref \undefined \def \botherref #1{#1}\fi
\ifx \url \undefined \def \url#1{\textsf{#1}}\fi
\ifx \bchapter \undefined \def \bchapter#1{#1}\fi
\ifx \bbook \undefined \def \bbook#1{#1}\fi
\ifx \bcomment \undefined \def \bcomment#1{#1}\fi
\ifx \oauthor \undefined \def \oauthor#1{#1}\fi
\ifx \citeauthoryear \undefined \def \citeauthoryear#1{#1}\fi
\ifx \endbibitem  \undefined \def \endbibitem {}\fi
\ifx \bconflocation  \undefined \def \bconflocation#1{#1}\fi
\ifx \arxivurl  \undefined \def \arxivurl#1{\textsf{#1}}\fi
\csname PreBibitemsHook\endcsname

%%% 1
\bibitem{Witkos2011}
\begin{barticle}
\bauthor{\bsnm{Witkos}, \binits{T.M.}},
\bauthor{\bsnm{Koscianska}, \binits{E.}},
\bauthor{\bsnm{Krzyzosiak}, \binits{W.J.}}:
\batitle{Practical aspects of microrna target prediction}.
\bjtitle{Curr Mol Med}
\bvolume{11}(\bissue{2}),
\bfpage{99}--\blpage{109}
(\byear{2011}).
doi:\doiurl{10.2174/156652411794859250}
\end{barticle}
\endbibitem

%%% 2
\bibitem{Ross2007}
\begin{barticle}
\bauthor{\bsnm{Ross}, \binits{J.S.}},
\bauthor{\bsnm{Carlson}, \binits{J.A.}},
\bauthor{\bsnm{Brock}, \binits{G.}}:
\batitle{mirna: the new gene silencer}.
\bjtitle{Am J Clin Pathol.}
\bvolume{128}(\bissue{5}),
\bfpage{830}--\blpage{836}
(\byear{2007}).
doi:\doiurl{10.1309/2JK279BU2G743MWJ}
\end{barticle}
\endbibitem

%%% 3
\bibitem{Chen2005}
\begin{barticle}
\bauthor{\bsnm{Chen}, \binits{P.Y.}},
\bauthor{\bsnm{Manninga}, \binits{H.}},
\bauthor{\bsnm{Slanchev}, \binits{K.}},
\bauthor{\bsnm{Chien}, \binits{M.}},
\bauthor{\bsnm{Russo}, \binits{J.J.}},
\bauthor{\bsnm{Ju}, \binits{J.}},
\bauthor{\bsnm{Sheridan}, \binits{R.}},
\bauthor{\bsnm{John}, \binits{B.}},
\bauthor{\bsnm{Marks}, \binits{D.S.}},
\bauthor{\bsnm{Gaidatzis}, \binits{D.}},
\bauthor{\bsnm{Sander}, \binits{C.}},
\bauthor{\bsnm{Zavolan}, \binits{M.}},
\bauthor{\bsnm{Tuschl}, \binits{T.}}:
\batitle{The developmental mirna profiles of zebrafish as determined by small
  rna cloning}.
\bjtitle{Genes and Development}
\bvolume{19}(\bissue{11}),
\bfpage{1288}--\blpage{1293}
(\byear{2005}).
doi:\doiurl{10.1101/gad.1310605}
\end{barticle}
\endbibitem

%%% 4
\bibitem{Xue2005}
\begin{barticle}
\bauthor{\bsnm{Xue}, \binits{C.}},
\bauthor{\bsnm{Li}, \binits{F.}},
\bauthor{\bsnm{He}, \binits{T.}},
\bauthor{\bsnm{Liu}, \binits{G.P.}},
\bauthor{\bsnm{Li}, \binits{Y.}},
\bauthor{\bsnm{Zhang}, \binits{X.}}:
\batitle{Classification of real and pseudo microrna precursors using local
  structure sequence features and support vector machine}.
\bjtitle{BMC Bioinformatics}
\bvolume{6},
\bfpage{310}
(\byear{2005}).
doi:\doiurl{10.1186/1471-2105-6-310}
\end{barticle}
\endbibitem

%%% 5
\bibitem{Huang2007}
\begin{barticle}
\bauthor{\bsnm{Huang}, \binits{T.H.}},
\bauthor{\bsnm{Fan}, \binits{B.}},
\bauthor{\bsnm{Rothschild}, \binits{M.F.}},
\bauthor{\bsnm{Hu}, \binits{Z.L.}},
\bauthor{\bsnm{Li}, \binits{K.}},
\bauthor{\bsnm{Zhao}, \binits{S.H.}}:
\batitle{Mirfinder: an improved approach and software implementation for
  genome-wide fast microrna precursor scans}.
\bjtitle{BMC Bioinformatics}
\bvolume{8},
\bfpage{341}
(\byear{2007}).
doi:\doiurl{10.1186/1471-2105-8-341}
\end{barticle}
\endbibitem

%%% 6
\bibitem{Ng2007}
\begin{barticle}
\bauthor{\bsnm{Ng}, \binits{K.L.S.}},
\bauthor{\bsnm{Mishra}, \binits{S.K.}}:
\batitle{De novo svm classification of precursor micrornas from genomic pseudo
  hairpins using global and intrinsic folding measures}.
\bjtitle{BMC Bioinformatics}
\bvolume{23}(\bissue{11}),
\bfpage{1321}--\blpage{1330}
(\byear{2007}).
doi:\doiurl{10.1186/1471-2105-8-341}
\end{barticle}
\endbibitem

%%% 7
\bibitem{Batuwita2009}
\begin{barticle}
\bauthor{\bsnm{Batuwita}, \binits{R.}},
\bauthor{\bsnm{Palade}, \binits{V.}}:
\batitle{micropred: effective classification of pre-mirnas for human mirna gene
  prediction}.
\bjtitle{BMC Bioinformatics}
\bvolume{25}(\bissue{8}),
\bfpage{989}--\blpage{995}
(\byear{2009}).
doi:\doiurl{10.1093/bioinformatics/btp107}
\end{barticle}
\endbibitem

%%% 8
\bibitem{Pasaila2011}
\begin{barticle}
\bauthor{\bsnm{Pasaila}, \binits{D.}},
\bauthor{\bsnm{Sucial}, \binits{A.}},
\bauthor{\bsnm{Mohorianu}, \binits{I.}},
\bauthor{\bsnm{Pantiru}, \binits{S.T.}},
\bauthor{\bsnm{Ciortuz}, \binits{L.}}:
\batitle{Mirna recognition with the yasmir system: The quest for further
  improvements}.
\bjtitle{Adv Exp Med Biol.}
\bvolume{696},
\bfpage{17}--\blpage{25}
(\byear{2011}).
doi:\doiurl{10.1007/978-1-4419-7046-6 2}
\end{barticle}
\endbibitem

%%% 9
\bibitem{Ding2010}
\begin{barticle}
\bauthor{\bsnm{Ding}, \binits{J.}},
\bauthor{\bsnm{Zhou}, \binits{S.}},
\bauthor{\bsnm{Guan}, \binits{J.}}:
\batitle{Mirensvm: towards better prediction of microrna precursors using an
  ensemble svm classifier with multi loop features}.
\bjtitle{BMC Bioinformatics}
\bvolume{14}(\bissue{11}),
\bfpage{11}--\blpage{11}
(\byear{2010}).
doi:\doiurl{10.1186/1471-2105-11-S11-S11}
\end{barticle}
\endbibitem

%%% 10
\bibitem{Wu2011}
\begin{barticle}
\bauthor{\bsnm{Wu}, \binits{Y.}},
\bauthor{\bsnm{Wei}, \binits{B.}},
\bauthor{\bsnm{Liu}, \binits{H.}},
\bauthor{\bsnm{Li}, \binits{T.}},
\bauthor{\bsnm{Rayner}, \binits{S.}}:
\batitle{Mirpara: a svm-based software tool for prediction of most probable
  microrna coding regions in genome scale sequences}.
\bjtitle{BMC Bioinformatics}
\bvolume{12},
\bfpage{107}
(\byear{2011}).
doi:\doiurl{10.1186/1471-2105-12-107}
\end{barticle}
\endbibitem

%%% 11
\bibitem{Kleftogiannis2015}
\begin{barticle}
\bauthor{\bsnm{Kleftogiannis}, \binits{D.}},
\bauthor{\bsnm{Theofilatos}, \binits{K.}},
\bauthor{\bsnm{Likothanassis}, \binits{S.}},
\bauthor{\bsnm{Mavroudi}, \binits{S.}}:
\batitle{Yamipred: A novel evolutionary method for predicting pre-mirnas and
  selecting relevant features}.
\bjtitle{IEEE ACM Transactions on Computational Biology and Bioinformatics}
\bvolume{12}(\bissue{5}),
\bfpage{1183}--\blpage{1192}
(\byear{2015}).
doi:\doiurl{10.1109/TCBB.2014.2388227}
\end{barticle}
\endbibitem

%%% 12
\bibitem{Hsieh2010}
\begin{barticle}
\bauthor{\bsnm{Hsieh}, \binits{C.H.}},
\bauthor{\bsnm{Chang}, \binits{D.T.H.}},
\bauthor{\bsnm{Hsueh}, \binits{C.H.}},
\bauthor{\bsnm{Wu}, \binits{C.Y.}},
\bauthor{\bsnm{Oyang}, \binits{Y.J.}}:
\batitle{Predicting microrna precursors with a generalized gaussian components
  based density estimation algorithm}.
\bjtitle{BMC Bioinformatics}
\bvolume{11},
\bfpage{1}--\blpage{52}
(\byear{2010}).
doi:\doiurl{10.1186/1471-2105-11-S1-S52}
\end{barticle}
\endbibitem

%%% 13
\bibitem{Wang2011}
\begin{barticle}
\bauthor{\bsnm{Wang}, \binits{Y.}},
\bauthor{\bsnm{Chen}, \binits{X.}},
\bauthor{\bsnm{Jiang}, \binits{W.}},
\bauthor{\bsnm{Li}, \binits{L.}},
\bauthor{\bsnm{Li}, \binits{W.}},
\bauthor{\bsnm{Yang}, \binits{L.}},
\bauthor{\bsnm{Liao}, \binits{M.}},
\bauthor{\bsnm{Lian}, \binits{B.}},
\bauthor{\bsnm{Lv}, \binits{Y.}},
\bauthor{\bsnm{Wang}, \binits{S.}},
\bauthor{\bsnm{Wang}, \binits{S.}},
\bauthor{\bsnm{Li}, \binits{X.}}:
\batitle{Predicting human microrna precursors based on an optimized feature
  subset generated by ga-svm}.
\bjtitle{Genomics}
\bvolume{98}(\bissue{2}),
\bfpage{73}--\blpage{78}
(\byear{2011})
\end{barticle}
\endbibitem

%%% 14
\bibitem{Xiao2011}
\begin{botherref}
\oauthor{\bsnm{Xiao}, \binits{J.}},
\oauthor{\bsnm{Tang}, \binits{X.}},
\oauthor{\bsnm{Li}, \binits{Y.}},
\oauthor{\bsnm{Fang}, \binits{Z.}},
\oauthor{\bsnm{Ma}, \binits{D.}},
\oauthor{\bsnm{He}, \binits{Y.}},
\oauthor{\bsnm{Li}, \binits{M.}}:
Identification of microrna precursors based on random forest with network-level
  representation method of stem-loop structure.
BMC Bioinformatics
\textbf{12:165}
(2011).
doi:\doiurl{10.1186/1471-2105-12-165}
\end{botherref}
\endbibitem

%%% 15
\bibitem{Rahman2012}
\begin{barticle}
\bauthor{\bsnm{Rahman}, \binits{M.E.}},
\bauthor{\bsnm{Islam}, \binits{R.}},
\bauthor{\bsnm{Islam}, \binits{S.}},
\bauthor{\bsnm{Mondal}, \binits{S.I.}},
\bauthor{\bsnm{Amin}, \binits{M.R.l.}}:
\batitle{Mirann: A reliable approach for improved classification of precursor
  microrna using artificial neural network model}.
\bjtitle{Genomics}
\bvolume{99},
\bfpage{189}--\blpage{194}
(\byear{2012})
\end{barticle}
\endbibitem

%%% 16
\bibitem{Thomas2017}
\begin{bchapter}
\bauthor{\bsnm{Thomas}, \binits{J.}},
\bauthor{\bsnm{Thomas}, \binits{S.}},
\bauthor{\bsnm{Sael}, \binits{L.}}:
\bctitle{{DP-miRNA: an improved prediction of precursor microRNA using deep
  learning mode}}.
In: \bbtitle{IEEE International Conference on Big Data and Smart Computing
  (IEEE BigComp 2017)},
pp. \bfpage{96}--\blpage{99}
(\byear{2017}).
\burl{http://conf2017.bigcomputing.org/}
\end{bchapter}
\endbibitem

%%% 17
\bibitem{Hinton2002}
\begin{barticle}
\bauthor{\bsnm{Hinton}, \binits{G.E.}}:
\batitle{{Training Products of Experts by Minimizing Contrastive Divergence}}.
\bjtitle{Neural Computation}
\bvolume{14}(\bissue{8}),
\bfpage{1771}--\blpage{1800}
(\byear{2002})
\end{barticle}
\endbibitem

%%% 18
\bibitem{Hinton2012}
\begin{botherref}
\oauthor{\bsnm{Hinton}, \binits{G.}},
\oauthor{\bsnm{Deng}, \binits{L.}},
\oauthor{\bsnm{Yu}, \binits{D.}},
\oauthor{\bsnm{Dahl}, \binits{G.}},
\oauthor{\bsnm{Mohamed}, \binits{A.R.}},
\oauthor{\bsnm{Jaitly}, \binits{N.}},
\oauthor{\bsnm{Vanhoucke}, \binits{V.}},
\oauthor{\bsnm{Nguyen}, \binits{P.}},
\oauthor{\bsnm{Sainath}, \binits{T.}},
\oauthor{\bsnm{Kingsbury}, \binits{B.}}:
Deep neural networks for acoustic modeling in speech recognition: The shared
  views of four research groups.
IEEE Signal Processing Magazine,
82--97
(2012).
doi:\doiurl{10.1.1.248.3619}
\end{botherref}
\endbibitem

%%% 19
\bibitem{Zhong2015}
\begin{barticle}
\bauthor{\bsnm{Zhong}, \binits{Y.}},
\bauthor{\bsnm{Xuan}, \binits{P.}},
\bauthor{\bsnm{Han}, \binits{K.}},
\bauthor{\bsnm{Zhang}, \binits{W.}},
\bauthor{\bsnm{Li}, \binits{J.}}:
\batitle{Improved pre-mirna classification by reducing the effect of class
  imbalance}.
\bjtitle{BioMed Research International}
\bvolume{2015},
\bfpage{1}--\blpage{12}
(\byear{2015}).
doi:\doiurl{10.1155/2015/960108}
\end{barticle}
\endbibitem

%%% 20
\bibitem{Yen2009}
\begin{barticle}
\bauthor{\bsnm{Yen}, \binits{S.J.}},
\bauthor{\bsnm{Lee}, \binits{Y.S.}}:
\batitle{Cluster-based under-sampling approaches for imbalanced data
  distributions}.
\bjtitle{Expert Systems with Applications}
\bvolume{36}(\bissue{3}),
\bfpage{5718}--\blpage{5727}
(\byear{2009}).
doi:\doiurl{10.1016/j.eswa.2008.06.108}
\end{barticle}
\endbibitem

\end{thebibliography}

\newcommand{\BMCxmlcomment}[1]{}

\BMCxmlcomment{

<refgrp>

<bibl id="B1">
  <title><p>Practical Aspects of microRNA Target Prediction</p></title>
  <aug>
    <au><snm>Witkos</snm><fnm>T. M</fnm></au>
    <au><snm>Koscianska</snm><fnm>E</fnm></au>
    <au><snm>Krzyzosiak</snm><fnm>W.J</fnm></au>
  </aug>
  <source>Curr Mol Med</source>
  <pubdate>2011</pubdate>
  <volume>11</volume>
  <issue>2</issue>
  <fpage>99</fpage>
  <lpage>-109</lpage>
</bibl>

<bibl id="B2">
  <title><p>miRNA: the new gene silencer</p></title>
  <aug>
    <au><snm>Ross</snm><fnm>J. S</fnm></au>
    <au><snm>Carlson</snm><fnm>J. A</fnm></au>
    <au><snm>Brock</snm><fnm>G.</fnm></au>
  </aug>
  <source>Am J Clin Pathol.</source>
  <pubdate>2007</pubdate>
  <volume>128</volume>
  <issue>5</issue>
  <fpage>830</fpage>
  <lpage>-836</lpage>
</bibl>

<bibl id="B3">
  <title><p>The developmental miRNA profiles of zebrafish as determined by
  small RNA cloning</p></title>
  <aug>
    <au><snm>Chen</snm><fnm>P. Y</fnm></au>
    <au><snm>Manninga</snm><fnm>H.</fnm></au>
    <au><snm>Slanchev</snm><fnm>K.</fnm></au>
    <au><snm>Chien</snm><fnm>M.</fnm></au>
    <au><snm>Russo</snm><fnm>J. J</fnm></au>
    <au><snm>Ju</snm><fnm>J.</fnm></au>
    <au><snm>Sheridan</snm><fnm>R.</fnm></au>
    <au><snm>John</snm><fnm>B.</fnm></au>
    <au><snm>Marks</snm><fnm>D. S</fnm></au>
    <au><snm>Gaidatzis</snm><fnm>D.</fnm></au>
    <au><snm>Sander</snm><fnm>C.</fnm></au>
    <au><snm>Zavolan</snm><fnm>M.</fnm></au>
    <au><snm>Tuschl</snm><fnm>T.</fnm></au>
  </aug>
  <source>Genes and Development</source>
  <pubdate>2005</pubdate>
  <volume>19</volume>
  <issue>11</issue>
  <fpage>1288</fpage>
  <lpage>-1293</lpage>
</bibl>

<bibl id="B4">
  <title><p>Classification of real and pseudo microRNA precursors using local
  Structure sequence features and support vector machine</p></title>
  <aug>
    <au><snm>Xue</snm><fnm>C.</fnm></au>
    <au><snm>Li</snm><fnm>F.</fnm></au>
    <au><snm>He</snm><fnm>T.</fnm></au>
    <au><snm>Liu</snm><fnm>G.P</fnm></au>
    <au><snm>Li</snm><fnm>Y.</fnm></au>
    <au><snm>Zhang</snm><fnm>X.</fnm></au>
  </aug>
  <source>BMC Bioinformatics</source>
  <pubdate>2005</pubdate>
  <volume>6</volume>
  <fpage>310</fpage>
</bibl>

<bibl id="B5">
  <title><p>MiRFinder: an improved approach and software implementation for
  genome-wide fast microRNA precursor scans</p></title>
  <aug>
    <au><snm>Huang</snm><fnm>T. H</fnm></au>
    <au><snm>Fan</snm><fnm>B.</fnm></au>
    <au><snm>Rothschild</snm><fnm>M. F</fnm></au>
    <au><snm>Hu</snm><fnm>Z. L</fnm></au>
    <au><snm>Li</snm><fnm>K.</fnm></au>
    <au><snm>Zhao</snm><fnm>S. H</fnm></au>
  </aug>
  <source>BMC Bioinformatics</source>
  <pubdate>2007</pubdate>
  <volume>8</volume>
  <fpage>341</fpage>
</bibl>

<bibl id="B6">
  <title><p>De novo SVM classification of precursor microRNAs from genomic
  pseudo hairpins using global and intrinsic folding measures</p></title>
  <aug>
    <au><snm>Ng</snm><fnm>K. L S</fnm></au>
    <au><snm>Mishra</snm><fnm>S. K</fnm></au>
  </aug>
  <source>BMC Bioinformatics</source>
  <pubdate>2007</pubdate>
  <volume>23</volume>
  <issue>11</issue>
  <fpage>1321</fpage>
  <lpage>-1330</lpage>
</bibl>

<bibl id="B7">
  <title><p>microPred: effective classification of pre-miRNAs for human miRNA
  gene prediction</p></title>
  <aug>
    <au><snm>Batuwita</snm><fnm>R.</fnm></au>
    <au><snm>Palade</snm><fnm>V.</fnm></au>
  </aug>
  <source>BMC Bioinformatics</source>
  <pubdate>2009</pubdate>
  <volume>25</volume>
  <issue>8</issue>
  <fpage>989</fpage>
  <lpage>-995</lpage>
</bibl>

<bibl id="B8">
  <title><p>MiRNA Recognition with the yasMiR System: The Quest for Further
  Improvements</p></title>
  <aug>
    <au><snm>Pasaila</snm><fnm>D.</fnm></au>
    <au><snm>Sucial</snm><fnm>A.</fnm></au>
    <au><snm>Mohorianu</snm><fnm>I.</fnm></au>
    <au><snm>Pantiru</snm><fnm>S. T</fnm></au>
    <au><snm>Ciortuz</snm><fnm>L.</fnm></au>
  </aug>
  <source>Adv Exp Med Biol.</source>
  <pubdate>2011</pubdate>
  <volume>696</volume>
  <fpage>17</fpage>
  <lpage>-25</lpage>
</bibl>

<bibl id="B9">
  <title><p>MiRenSVM: towards better prediction of microRNA precursors using an
  ensemble SVM classifier with multi loop features</p></title>
  <aug>
    <au><snm>Ding</snm><fnm>J.</fnm></au>
    <au><snm>Zhou</snm><fnm>S.</fnm></au>
    <au><snm>Guan</snm><fnm>J.</fnm></au>
  </aug>
  <source>BMC Bioinformatics</source>
  <pubdate>2010</pubdate>
  <volume>14</volume>
  <issue>11</issue>
  <fpage>Suppl11:S11</fpage>
</bibl>

<bibl id="B10">
  <title><p>MiRPara: a SVM-based software tool for prediction of most probable
  microRNA coding regions in genome scale sequences</p></title>
  <aug>
    <au><snm>Wu</snm><fnm>Y.</fnm></au>
    <au><snm>Wei</snm><fnm>B.</fnm></au>
    <au><snm>Liu</snm><fnm>H.</fnm></au>
    <au><snm>Li</snm><fnm>T.</fnm></au>
    <au><snm>Rayner</snm><fnm>S.</fnm></au>
  </aug>
  <source>BMC Bioinformatics</source>
  <pubdate>2011</pubdate>
  <volume>12</volume>
  <fpage>107</fpage>
</bibl>

<bibl id="B11">
  <title><p>YamiPred: A Novel Evolutionary Method for Predicting Pre-miRNAs and
  Selecting Relevant Features</p></title>
  <aug>
    <au><snm>Kleftogiannis</snm><fnm>D.</fnm></au>
    <au><snm>Theofilatos</snm><fnm>K.</fnm></au>
    <au><snm>Likothanassis</snm><fnm>S.</fnm></au>
    <au><snm>Mavroudi</snm><fnm>S.</fnm></au>
  </aug>
  <source>IEEE ACM Transactions on Computational Biology and
  Bioinformatics</source>
  <pubdate>2015</pubdate>
  <volume>12</volume>
  <issue>5</issue>
  <fpage>1183</fpage>
  <lpage>-1192</lpage>
</bibl>

<bibl id="B12">
  <title><p>Predicting microRNA precursors with a generalized Gaussian
  components based density estimation algorithm</p></title>
  <aug>
    <au><snm>Hsieh</snm><fnm>C.H.</fnm></au>
    <au><snm>Chang</snm><fnm>D. T. H.</fnm></au>
    <au><snm>Hsueh</snm><fnm>C. H.</fnm></au>
    <au><snm>Wu</snm><fnm>C. Y.</fnm></au>
    <au><snm>Oyang</snm><fnm>Y. J.</fnm></au>
  </aug>
  <source>BMC Bioinformatics</source>
  <pubdate>2010</pubdate>
  <volume>11</volume>
  <fpage>(Suppl1):S52</fpage>
</bibl>

<bibl id="B13">
  <title><p>Predicting human microRNA precursors based on an optimized feature
  subset generated by GA-SVM</p></title>
  <aug>
    <au><snm>Wang</snm><fnm>Y.</fnm></au>
    <au><snm>Chen</snm><fnm>X.</fnm></au>
    <au><snm>Jiang</snm><fnm>W.</fnm></au>
    <au><snm>Li</snm><fnm>L.</fnm></au>
    <au><snm>Li</snm><fnm>W.</fnm></au>
    <au><snm>Yang</snm><fnm>L.</fnm></au>
    <au><snm>Liao</snm><fnm>M.</fnm></au>
    <au><snm>Lian</snm><fnm>B.</fnm></au>
    <au><snm>Lv</snm><fnm>Y.</fnm></au>
    <au><snm>Wang</snm><fnm>S.</fnm></au>
    <au><snm>Wang</snm><fnm>S.</fnm></au>
    <au><snm>Li</snm><fnm>X.</fnm></au>
  </aug>
  <source>Genomics</source>
  <pubdate>2011</pubdate>
  <volume>98</volume>
  <issue>2</issue>
  <fpage>73</fpage>
  <lpage>-78</lpage>
</bibl>

<bibl id="B14">
  <title><p>Identification of microRNA precursors based on random forest with
  network-level representation method of stem-loop structure</p></title>
  <aug>
    <au><snm>Xiao</snm><fnm>J.</fnm></au>
    <au><snm>Tang</snm><fnm>X.</fnm></au>
    <au><snm>Li</snm><fnm>Y.</fnm></au>
    <au><snm>Fang</snm><fnm>Z.</fnm></au>
    <au><snm>Ma</snm><fnm>D.</fnm></au>
    <au><snm>He</snm><fnm>Y.</fnm></au>
    <au><snm>Li</snm><fnm>M.</fnm></au>
  </aug>
  <source>BMC Bioinformatics</source>
  <pubdate>2011</pubdate>
  <volume>12:165</volume>
</bibl>

<bibl id="B15">
  <title><p>MiRANN: A reliable approach for improved classification of
  precursor microRNA using Artificial Neural Network model</p></title>
  <aug>
    <au><snm>Rahman</snm><fnm>M. E</fnm></au>
    <au><snm>Islam</snm><fnm>R.</fnm></au>
    <au><snm>Islam</snm><fnm>S.</fnm></au>
    <au><snm>Mondal</snm><fnm>S. I</fnm></au>
    <au><snm>Amin</snm><fnm>MR</fnm></au>
  </aug>
  <source>Genomics</source>
  <pubdate>2012</pubdate>
  <volume>99</volume>
  <fpage>189</fpage>
  <lpage>-194</lpage>
</bibl>

<bibl id="B16">
  <title><p>{DP-miRNA: an improved prediction of precursor microRNA using deep
  learning mode}</p></title>
  <aug>
    <au><snm>Thomas</snm><fnm>J</fnm></au>
    <au><snm>Thomas</snm><fnm>S</fnm></au>
    <au><snm>Sael</snm><fnm>L</fnm></au>
  </aug>
  <source>IEEE International Conference on Big Data and Smart Computing (IEEE
  BigComp 2017)</source>
  <pubdate>2017</pubdate>
  <fpage>96</fpage>
  <lpage>-99</lpage>
  <url>http://conf2017.bigcomputing.org/</url>
</bibl>

<bibl id="B17">
  <title><p>{Training Products of Experts by Minimizing Contrastive
  Divergence}</p></title>
  <aug>
    <au><snm>Hinton</snm><fnm>GE</fnm></au>
  </aug>
  <source>Neural Computation</source>
  <pubdate>2002</pubdate>
  <volume>14</volume>
  <issue>8</issue>
  <fpage>1771</fpage>
  <lpage>-1800</lpage>
</bibl>

<bibl id="B18">
  <title><p>Deep Neural Networks for Acoustic Modeling in Speech Recognition:
  The shared views of four research groups</p></title>
  <aug>
    <au><snm>Hinton</snm><fnm>G.</fnm></au>
    <au><snm>Deng</snm><fnm>L.</fnm></au>
    <au><snm>Yu</snm><fnm>D.</fnm></au>
    <au><snm>Dahl</snm><fnm>G.</fnm></au>
    <au><snm>Mohamed</snm><fnm>A. R.</fnm></au>
    <au><snm>Jaitly</snm><fnm>N.</fnm></au>
    <au><snm>Vanhoucke</snm><fnm>V.</fnm></au>
    <au><snm>Nguyen</snm><fnm>P.</fnm></au>
    <au><snm>Sainath</snm><fnm>T.</fnm></au>
    <au><snm>Kingsbury</snm><fnm>B.</fnm></au>
  </aug>
  <source>IEEE Signal Processing Magazine</source>
  <pubdate>2012</pubdate>
  <fpage>82</fpage>
  <lpage>-97</lpage>
</bibl>

<bibl id="B19">
  <title><p>Improved Pre-miRNA Classification by Reducing the Effect of Class
  Imbalance</p></title>
  <aug>
    <au><snm>Zhong</snm><fnm>Y.</fnm></au>
    <au><snm>Xuan</snm><fnm>P.</fnm></au>
    <au><snm>Han</snm><fnm>K.</fnm></au>
    <au><snm>Zhang</snm><fnm>W.</fnm></au>
    <au><snm>Li</snm><fnm>J.</fnm></au>
  </aug>
  <source>BioMed Research International</source>
  <pubdate>2015</pubdate>
  <volume>2015</volume>
  <fpage>1</fpage>
  <lpage>-12</lpage>
</bibl>

<bibl id="B20">
  <title><p>Cluster-based under-sampling approaches for imbalanced data
  distributions</p></title>
  <aug>
    <au><snm>Yen</snm><fnm>S. J</fnm></au>
    <au><snm>Lee</snm><fnm>Y. S</fnm></au>
  </aug>
  <source>Expert Systems with Applications</source>
  <pubdate>2009</pubdate>
  <volume>36</volume>
  <issue>3</issue>
  <fpage>5718</fpage>
  <lpage>-5727</lpage>
</bibl>

</refgrp>
} % end of \BMCxmlcomment
% for author-year bibliography (bmc-mathphys or spbasic)
% a) write to bib file (bmc-mathphys only)
% @settings{label, options="nameyear"}

% or include bibliography directly:
% \begin{thebibliography}
% \bibitem{b1}
% \end{thebibliography}

%%%%%%%%%%%%%%%%%%%%%%%%%%%%%%%%%%%
%%                               %%
%% Figures                       %%
%%                               %%
%% NB: this is for captions and  %%
%% Titles. All graphics must be  %%
%% submitted separately and NOT  %%
%% included in the Tex document  %%
%%                               %%
%%%%%%%%%%%%%%%%%%%%%%%%%%%%%%%%%%%

%%
%% Do not use \listoffigures as most will included as separate files

\section*{Figures}
\begin{comment}
  \begin{figure}[h!]
  \caption{\csentence{Sample figure title.}
      A short description of the figure content
      should go here.}
      \end{figure}

\begin{figure}[h!]
  \caption{\csentence{Sample figure title.}
      Figure legend text.}
      \end{figure}
\end{comment}

 \begin{figure}[h!]
\includegraphics[width=0.8\textwidth]{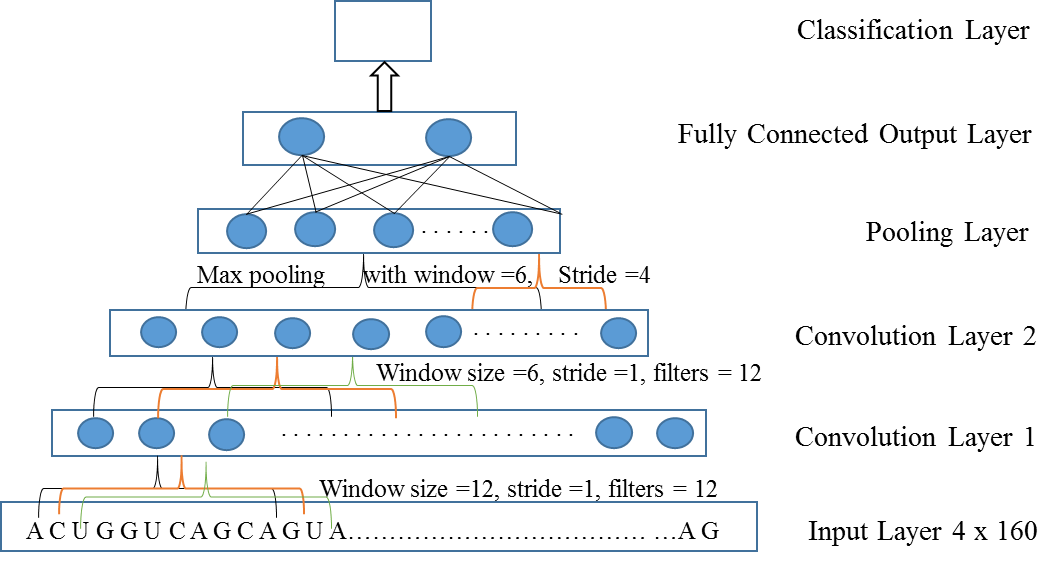}
  \caption{\csentence{Best performing CNN architecture.}
     }
 \label{fig:cnn}
\end{figure}

\begin{figure}[h!]
\includegraphics[width=0.8\textwidth]{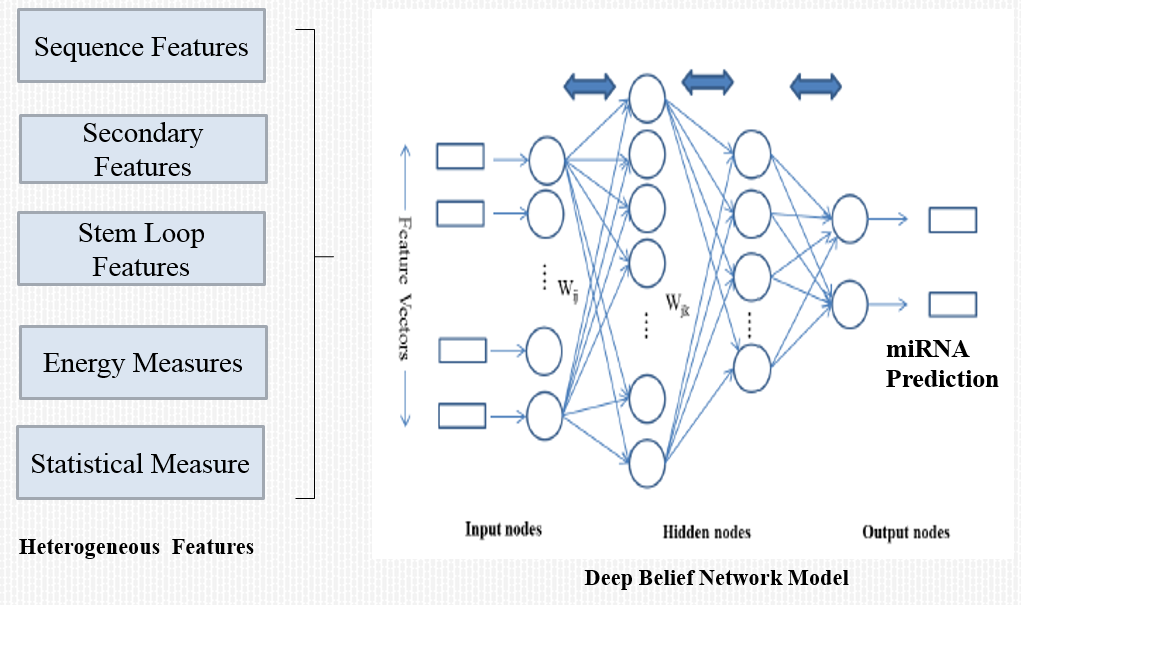}
  \caption{\csentence{Best performing DBN architecture.}
      }
  \label{fig:dbn}
\end{figure}

%%%%%%%%%%%%%%%%%%%%%%%%%%%%%%%%%%%
%%                               %%
%% Tables                        %%
%%                               %%
%%%%%%%%%%%%%%%%%%%%%%%%%%%%%%%%%%%

%% Use of \listoftables is discouraged.
%%
\section*{Tables}
\begin{table}[h!]
\caption{Sample table title. This is where the description of the table should go.}
      \begin{tabular}{cccc}
        \hline
           & B1  &B2   & B3\\ \hline
        A1 & 0.1 & 0.2 & 0.3\\
        A2 & ... & ..  & .\\
        A3 & ..  & .   & .\\ \hline
      \end{tabular}
\end{table}

\begin{comment}
\begin{table}
\begin{table*}[h!]
\caption{Class description for different species}
\label{table:00}
\centering
\hspace*{-3em}
%\begin{tabular}{p{1cm}p{0.9cm}p{1cm}p{0.7cm}p{0.7cm}p{0.7cm}p{0.4cm}p{1.1cm}p{0.6cm}p{0.7cm}p{1.4cm}p{0.7cm}}
 \begin{tabular}{llllllllllll}
    \hline
   % \multirow{2}{*}{Class} &
      \multicolumn{11}{c}{Species} & \\
            Class &Human&Caballus&Canis&Gallus&Gorilla&Mus&Panciscus&Pongo&Taurus&Troglodytes&Virus\\
  \hline
Positive&	1600&341&323&497&85&720&88&581&662&599&237 \\
Negative&	8494&517&520&2000&1000&3000&116&1356&3000&176&107\\

\hline
\end{tabular}
\end{table*}
\end{comment}

\begin{table}[h!]
\centering
\caption{My caption}
\label{my-label}
\begin{tabular}{l p{8cm}}
\hline
Category & Features \\
\hline
Sequence composition properties & features related to the frequency of two and three adjacent nucleotide,                                                                                                                                                      aggregate dinucleotide frequency in the sequence, such as Dinucleotide pair frequency, Trinucleotide frequency, aggregate dinucleotide frequency \\
Secondary structures & thermodynamic stability profiles of pre-miRNAs \\
Stem and loop & diversity, frequency, entropy-related properties, enthalpy-related properties of the structure, hairpin length, loop length, consecutive base-pair, ratio of loop length to hairpin length of pre-miRNA secondary structure \\
Energy  characteristics & minimal free energy of the secondary structure, overall free energy NEFE, combined energy features, the energy required for dissolving the secondary structure \\
Statistical measures & Z-score of the folding measures zG, zQ, zSP, zP, zD \\
\hline
\end{tabular}
\end{table}

\begin{table}[h!]
\centering
\caption{Tested CNN architectures}
\label{tab:CNN_arc}
\begin{tabular}{l p{1.5cm} p{8cm}}
\hline
Model   Name                 & Number of Layers & Description of architecture   \\                                                                                                                                                                                    \hline
Type 1                 & 5                & Layer 1: Input layer, Layer 2: Convolution with no stride, Layer 3 Pooling layer with stride, Layer 4 Fully connected to output layer, Layer 5 classification layer                                               \\
Type 2                 & 5                & Layer 1: Input layer, Layer 2: Convolution with stride, Layer 3 Fully connected Layer , Layer 4 Fully connected to output layer, Layer 5 classification layer                                                     \\
Type 3                 & 6                & Layer 1: Input layer, Layer 2: Convolution with no stride,  Layer 3: Convolution with no stride, Layer 4 Pooling layer with stride, Layer 5 Fully connected to output layer, Layer 6 classification layer         \\
Type 4                 & 7                & Layer 1: Input layer, Layer 2: Convolution with no stride,  Layer 3: Convolution with no stride, Layer 4: Convolution with no stride, Layer 5 Pooling layer with stride, Layer 6 Fully connected to output layer, Layer 7 classification layer \\
\hline
\end{tabular}
\end{table}

\begin{table}[h!]
\centering
\caption{My caption}
\label{tab:CNN_hp}
\begin{tabular}{ll}
\hline
Hyper-parameter              & Range              \\
\hline
Filter size                  & 5 to 24            \\
Number of filters            & 5 to 20            \\
Stride                       & 0 to 24            \\
Pooling                      & Max pooling 0 to 9 \\
Dropout                      & 0 to 0.4           \\
Number of convolution layers & 1 to 3               \\
\hline
\end{tabular}
\end{table}

\begin{table}[h!]
\centering
\caption{Two selected CNN model description.}
\label{tab:CNN_best}
\begin{tabular}{lp{1.5cm}p{5cm}l}
\hline
Model Type & Description of model & & Performance Measure \\
\hline
\multirow{5}{*}{Type 2} & Layer 1 & Input Sequence & SE=1 \\
 & Layer 2 & Convolution Layer, Window size= 18, Stride = 4, & Number of filters =20. \\
 & Layer 3 & Fully connected Layer (90 neurons) & Precision= 0.985 \\
 & Layer 4 & Fully connected Layer (2 neurons) & Acc=0.993 \\
 & Layer 5 & Classification Layer &  \\
\hline
\multirow{6}{*}{Type 3} & Layer 1 & Input Sequence & SE=1 \\
 & Layer 2 & Convolution Layer (window size=12, stride=1, Num. of filters=12) & SP=0.990 \\
 & Layer 3 & Convolution Layer (window size=6, stride=1, Num. of filters=12) & Precision=0.990 \\
 & Layer 4 & Pooling Layer (max pooling, stride=4) & Acc=0.995  \\
 & Layer 5 & Fully Layer (2 neurons) & \\
 & Layer 6 & Classification Layer & \\
 \hline
\end{tabular}
\end{table}

\begin{table}[h!]
\centering
\caption{Comparison with existing computational intelligence techniques}
\label{tab:acc}
\begin{tabular}{llll}
\hline
Method                           & Sensitivity    & Specificity    & Accuracy \\
\hline
Naive Bayes                      & 0.943 & 0.796 & 0.914    \\
K nearest neighbors              & 0.970 & 0.657 & 0.908    \\
Random Forest                    & 0.979 & 0.765 & 0.937    \\
miRNN                            & 0.963 & 0.705 & 0.917    \\
YamiPred                         & 0.937 & 0.912 & 0.932    \\
Deep RBM model {[}58 features{]} & 0.973 & 0.942 & 0.968    \\
Deep RBM model {[}20 features{]} & 0.995 & 0.982 & 0.990    \\
CNN model 1 (Type 2)             & 1.00  & 0.985 & 0.993    \\
CNN model 2 (Type 3)             & 1.00  & 0.990 & 0.995    \\
\hline
\end{tabular}
\end{table}

%%%%%%%%%%%%%%%%%%%%%%%%%%%%%%%%%%%
%%                               %%
%% Additional Files              %%
%%                               %%
%%%%%%%%%%%%%%%%%%%%%%%%%%%%%%%%%%%

\section*{Additional Files}
\subsection*{Full list of pre-miRNA features}
Full list of pre-miRNA features used as inputs to deep belief network is listed.

\begin{table*}
\centering
\caption{Features for miRNA Prediction}
\label{tab:S1}
%\resizebox{1.3\columnwidth}{!}{%
 \begin{tabular}{p{2.5cm} c p{8cm}}
\hline
Feature&Number&Description\\
\hline
XY, where X,Y$\in$ $\{$A,C,G,U$\}$&16&	Dinucleotide pairs frequency\\
XYZ, where X,Y,Z$\in$ $\{$A,C,G,U$\}$&64&	Trinucleotide pairs frequency\\
A+U$\%$&1&Aggregate dinucleotide frequency (bases which are either A or U)\\
G+C$\%$&1&Aggregate dinucleotide frequency (bases which are either G or C)\\
L&1& Structure length\\
Freq&1&Structural frequency property \\
dP&1&Adjusted base pairing propensity given as  total$\_$bases$/$L\\
dG&1&Adjusted Minimum Free Energy of folding given as dG = MFE$/$L\\
dD&1&Adjusted base pair distance \\
dQ&1& Adjusted shannon entropy \\
dF&1&Compactness of the tree-graph representation of the sequence\\
MFEI1&1&MFEI1 = dG$/$$\%$(C+G)\\
MFEI2&1&MFEI2 = dG$/$number$\_$of$\_$stems\\
MFEI3&1&MFEI3 = dG$/$number$\_$of$\_$loops\\
MFEI4&1&MFEI4 = dG$/$total$\_$bases\\
MFEI5&1& MFEI5= dG$/$$\%$(A+U) ratio\\
 dS&1&Structure entropy\\
 dS$/$L&1&Normalized structure entropy\\
dH&1&Structure Enthalpy \\
dH$/$L&1&Normalized structure enthalpy \\
Tm&1& Melting Temperature \\
Tm$/$L&1&Normalized melting temperature \\
BP$/$X, where X $\in$ $\{$GC,GU,AU$\}$&3& Ratio of total$\_$bases to respective base pairs\\
 G$/$C &1&Number of G,C bases\\
Avg$\_$BP$\_$Stem&1&Average number of base pairs in the stem region \\
$\mid$A$-$U$\mid$$/$L,$\mid$G$-$C$\mid$$/$L, $\mid$G$-$U$\mid$$/$L&3& $\mid$X$-$Y$\mid$ is the number of (X $-$Y) base pairs in the secondary structure\\
$\mid$A$-$U$\mid$$\%$$/$n$\_$stems, $\mid$G $-$ C$\mid$$\%$$/$n$\_$stems, \\and $\mid$G $-$ U$\mid$$\%$$/$n$\_$stems
&3&Average number of base pairs in the stem region\\
zP,zG,zD,zQ,zSP&5&statistical Z-score of the folding measures\\
dPs &1& Positional Entropy which estimates the structural volatility of the secondary structure\\
EAFE&1&Normalized Ensemble Free Energy\\
CE$/$L&1& Centroid energy normalized by length\\
Diff&1& Diff =$\mid$ MFE-EFE$\mid$$/$L where, EFE is the ensemble free energy\\
IH&1&Hairpin length dangling ends\\
IL&1& Loop length\\
IC&1& Maximum consecutive base-pairs\\
$\%$L&1&Ratio of loop length to hairpin length\\
\hline
\end{tabular}
\end{table*}

\end{backmatter}
\end{document}